\documentclass[runningheads]{llncs}
\usepackage[T1]{fontenc}
\usepackage{graphicx}
\usepackage{amsmath} 
\usepackage{amssymb}
\usepackage{colortbl}
\usepackage{multirow}
\usepackage{color}
\usepackage{tabularray}
\usepackage{adjustbox}
\usepackage{hhline}
\usepackage{colortbl}
\usepackage{color}
\usepackage[misc]{ifsym}

\usepackage{hyperref}
\hypersetup{
    colorlinks=true,
    linkcolor=blue,
    filecolor=magenta,      
    urlcolor=blue,
    pdftitle={Overleaf Example},
    pdfpagemode=FullScreen,
    }

\newcommand{\repeatthanks}{\textsuperscript{\thefootnote}}

\usepackage[dvipsnames]{xcolor}
\begin{document}
%
\title{Instrument-Splatting: Controllable Photorealistic Reconstruction of Surgical Instruments Using Gaussian Splatting}
\titlerunning{Controllable Reconstruction of Surgical Instruments}
%
\author{Shuojue Yang\inst{1}\thanks{Equal contribution} \and Zijian Wu\inst{2}\repeatthanks \and
Mingxuan Hong\inst{1} \and
Qian Li\inst{1} \and
Daiyun Shen\inst{1}\ \and
Septimiu E. Salcudean\inst{2} \and
Yueming Jin\inst{1}\textsuperscript{(\Letter)}
}
%
\authorrunning{Anonymous Author et al.}
%
\institute{National University of Singapore, Singapore, Singapore \\
\email{ymjin@nus.edu.sg}\\
\and
The University of British Columbia, Vancouver, Canada}

\maketitle              
\begin{abstract}
Real2Sim is becoming increasingly important with the rapid development of surgical artificial intelligence (AI) and autonomy. In this work, we propose a novel Real2Sim methodology, \textit{Instrument-Splatting}, that leverages 3D Gaussian Splatting to provide fully controllable 3D reconstruction of surgical instruments from monocular surgical videos. To maintain both high visual fidelity and manipulability, we introduce a geometry pre-training to bind Gaussian point clouds on part mesh with accurate geometric priors and define a forward kinematics to control the Gaussians as flexible as real instruments. Afterward, to handle unposed videos, we design a novel instrument pose tracking method leveraging semantics-embedded Gaussians to robustly refine per-frame instrument poses and joint states in a render-and-compare manner, which allows our instrument Gaussian to accurately learn textures and reach photorealistic rendering. 
We validated our method on 2 publicly released surgical videos and 4 videos collected on \textit{ex vivo} tissues and green screens. 
Quantitative and qualitative evaluations demonstrate the effectiveness and superiority of the proposed method. 
\end{abstract}

\section{Introduction}

In robot-assisted laparoscopic surgery, computer vision-based surgical instrument identification~\cite{allan20183,chen2024mfst,du2018articulated,wu2024augmenting,zhan2024tracking,ye2016real} is of fundamental significance in downstream applications such as augmented reality and autonomous robotic surgery. 
While deep learning-based methods demonstrated excellent performance in general computer vision domains, adapting these methods to surgical instrument analysis remains challenging due to limited data and annotations.

For robotic surgery automation, existing public datasets either lack instrument pose labels or only provide sole wrist part pose labels~\cite{xu2025surgripe}. This limits imitation learning-based methods where the robot directly learns the instrumentation motions from the human expert surgical videos
~\cite{kim2024surgical,reiley2010motion}.  
To address this issue, a feasible solution is Sim2Real transfer learning, which leverages large amounts of trajectories and videos generated from the simulation environment~\cite{munawar2022open}. 
Conventionally, most studies~\cite{barragan2024realistic} adopt CAD mesh models of instruments that exhibit visually unrealistic appearances. This significant sim-to-real gap impairs the performance of models trained on such data. Therefore, a highly realistic and controllable 3D asset of surgical instruments is critical. 

With the recent progress in 3D reconstruction, new 3D representations such as Neural Radiance Field (NeRF)~\cite{mildenhall2021nerf} and 3D Gaussian Splatting (GS)~\cite{kerbl20233d} emerged, which serve as powerful tools that can provide reconstructions with high visual fidelity. Driven by these advances, Zeng et al.~\cite{zeng2024realistic} utilize 3D GS to reconstruct surgical instruments for novel data synthesis assisting surgical instrument detection. However, this method requires a customized system for data collection and cannot dynamically model surgical instruments with joint changes, which degrades the flexibility and controllability of the reconstructed 3D asset. Although previous methods~\cite{liu2024endoG,yang2024deform3dgs,liu2024lgs,zhu2024deformable,wang2022neural,yang2023neural} for dynamic surgical scene reconstruction have demonstrated high visual fidelity on deformable tissue reconstruction, they cannot provide controllable 3D GS since the deformation fields are only conditioned on the timestamps in surgical videos. 
In addition, the reconstruction quality of these methods is degraded with large and complex instrument motions, which commonly occur in surgical scenarios. 
Meanwhile, there are emerging works that adopt GS to reconstruct the general robotic arm~\cite{lou2024robo,lu2024manigaussian} or articulated human body~\cite{Kocabas_2024_CVPR}, these approaches assume known (or well-estimated) poses and joint states of the articulated objects, which is unpractical in robotic-surgery settings where the kinematics of surgical instruments are unavailable/noisy~\cite{allan20183}.

Driven by these limitations, we propose \textit{Instrument-Splatting}, a holistic framework, to reconstruct photorealistic surgical instruments represented by controllable 3D GS from untextured instrument CAD models and a monocular surgical video. The pipeline starts with a geometry pretraining which effectively binds the GS to the mesh models with accurate geometry. Next, we propose a novel pose estimation and tracking strategy based on the pretrained GS to estimate the instrument pose and joint states. To handle large inter-frame instrument motions, we first introduce a correspondence matching module to guide the GS to move toward the current frame pose, followed by a structural-prior-based loose regularization to penalize singular pose estimations. Finally, we evaluate our framework on both intraoperative videos selected from EndoVis2017 and EndoVis2018 and the in-house dataset collected using da Vinci robotic system. 
Experiments indicate the efficacy of our approach, demonstrating superior pose estimation accuracy and visual fidelity. 
\section{Method}
\begin{figure}[h]
\begin{minipage}{0.5\linewidth}
\centering
\includegraphics[width=6cm]{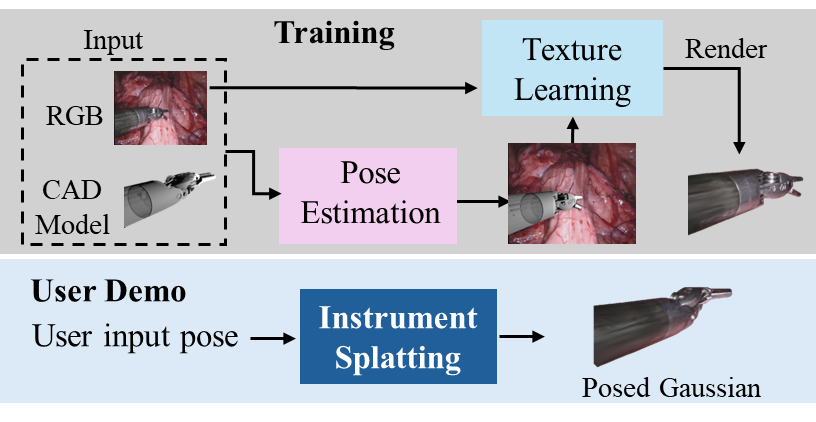}
\caption{Instrument-Splatting Pipeline.} \label{fig1}
\end{minipage}
\begin{minipage}{0.5\linewidth}
\centering
\includegraphics[width=6cm]{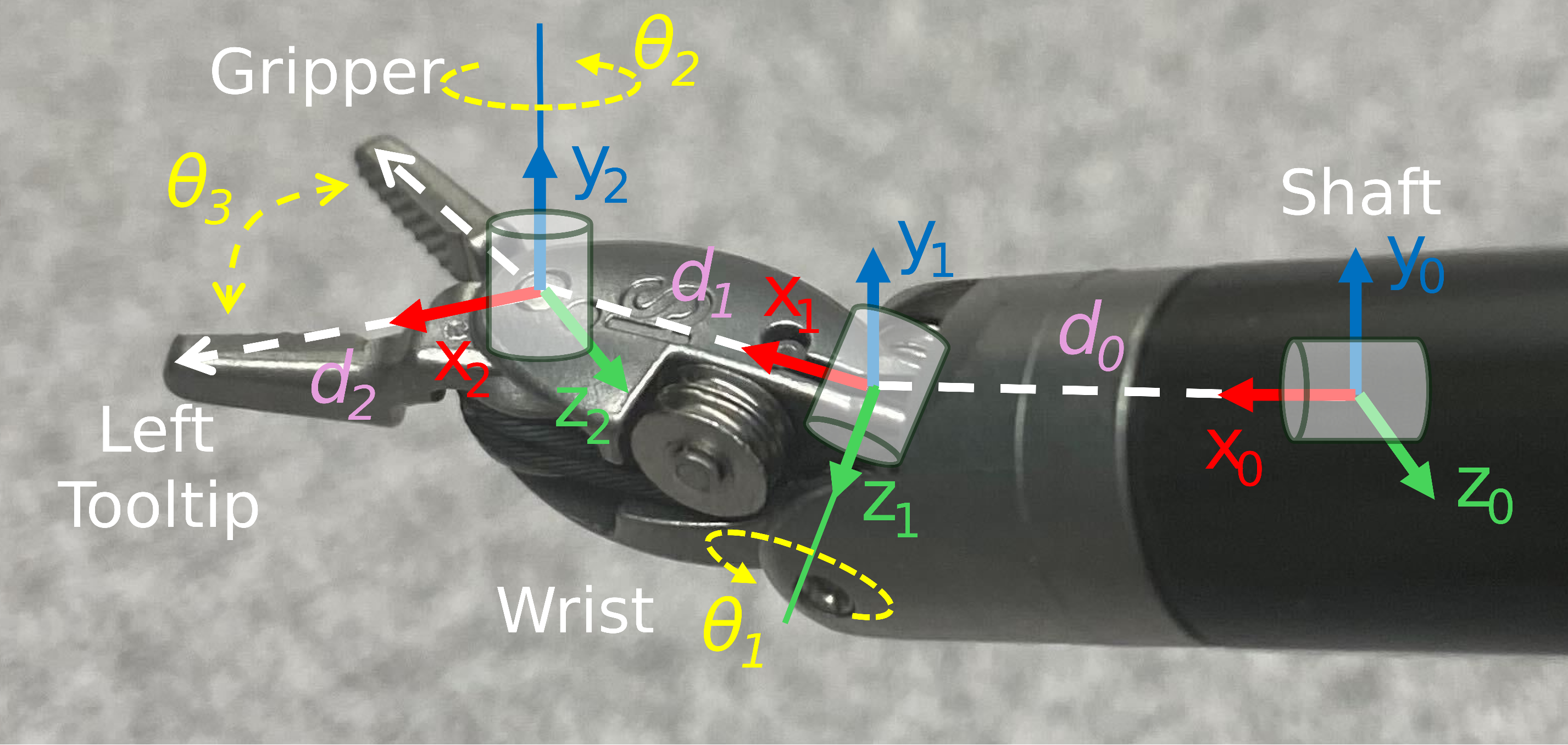}
\caption{Forward kinematics of the LND.} \label{fig2}
\end{minipage}
\end{figure}

Instrument-splatting aims to learn a Gaussian Splatting representation of articulated surgical instruments. The overview pipeline is shown in Fig.~\ref{fig1}. During training, we first estimate the instrument pose and joint state given an RGB image and instrument CAD model. Then, images with corresponding poses are fed into the texture learning module to optimize the instrument GS. After training, users can input in-ranged instrument pose with joint variables and get the posed mesh along with rendered photorealistic images. In the following, we present the preliminary knowledge including the instrument forward kinematics and 3D GS (Section ~\ref{2.1}), GS geometry pretraining (Section~\ref{2.2}), instrument pose estimation \& tracking (Section~\ref{2.3}), and the final texture learning (Section~\ref{2.4}).

\subsection{Preliminary Knowledge} \label{2.1}
\noindent\textbf{Forward Kinematics.}
In this study, we adopt the da Vinci EndoWrist Large Needle Driver (LND) as an example surgical instrument. The CAD model of this instrument is open source~\cite{kazanzides2014open}.
Note that the proposed method can be easily adapted to other types of instruments, given their CAD models. 

The definition of coordinate systems and forward kinematics diagram are shown in Fig.~\ref{fig2}. 
We regard shaft frame as the base frame and this frame definition consists of two revolute joint variables $\theta_1$ and $\theta_2$, denoting the rotation of the axis $z_1$ and $y_2$, and the gripper open angle $\theta_3$. Let $q=\{\theta_1,\;\theta_2,\;\theta_3\}\in\mathbb{R}^3$ represent the instrument joint state. Given the translation $d_0$ and $d_1$ as 215.9 mm and 9 mm, 
the forward kinematics chains can be defined as ${}^{s}T_w(\theta_1)\;{}^{w}T_{g}(\theta_2)\;{}^{g}T_{l}(\theta_3)$, where ${}^aT_b\in SE(3)$ denotes the rigid transformation from frame $a$ to $b$, and $s$, $w$, $g$ and $l$ denote the frames for shaft, wrist, gripper and left tool tips, respectively. Of note, each of them has only one degree of freedom (DoF) controlled by $\theta_i\in q$. Therefore, we can define a transformation from any joint frame $j$ to the shaft frame as ${}^s{T}_{j\in\{w,\;g\;,l\}}$ given corresponding joint states $q$. Next, with another transformation from shaft to camera ${}^{cam}T_s$, we define a transform function $\boldsymbol{T_j}(\cdot)$ to transform points $x_j\in\mathbb{R}^3$ in joint frames $j$ to the camera frame as:
\begin{equation}
\boldsymbol{T_j}(x_j;\;q,\xi)={}^{cam}T_s(\xi)\;{}^{s}T_j(q)\;x_j,
\label{eq:transformation}
\end{equation}
where we use $\xi\in se(3)$ to parameterize ${}^{cam}T_s$. With Eq.~\ref{eq:transformation}, we can define the instrument pose in camera frame given specified $(q,\;\xi)$. 
\noindent\textbf{3D Gaussian Splatting.}
3D Gaussian Splatting (GS)~\cite{kerbl20233d} is a powerful 3D representation that can reconstruct 3D scenes with a group of Gaussian points. Each Gaussian point contains learnable attributes $\Theta=\{{\mu}\in\mathbb{R}^{3},{r}\in\mathbb{R}^{4},{s}\in\mathbb{R}^{3},\alpha\in\mathbb{R},sh\in\mathbb{R}^{27}\}$ to define the position, rotation, scale, opacity and colors.   With camera parameters, $\alpha$-blending~\cite{kerbl20233d} is performed to render the colored images based on the attributes in 3D GS model.

\begin{figure}[t]
\centering
\includegraphics[width=0.8\textwidth]{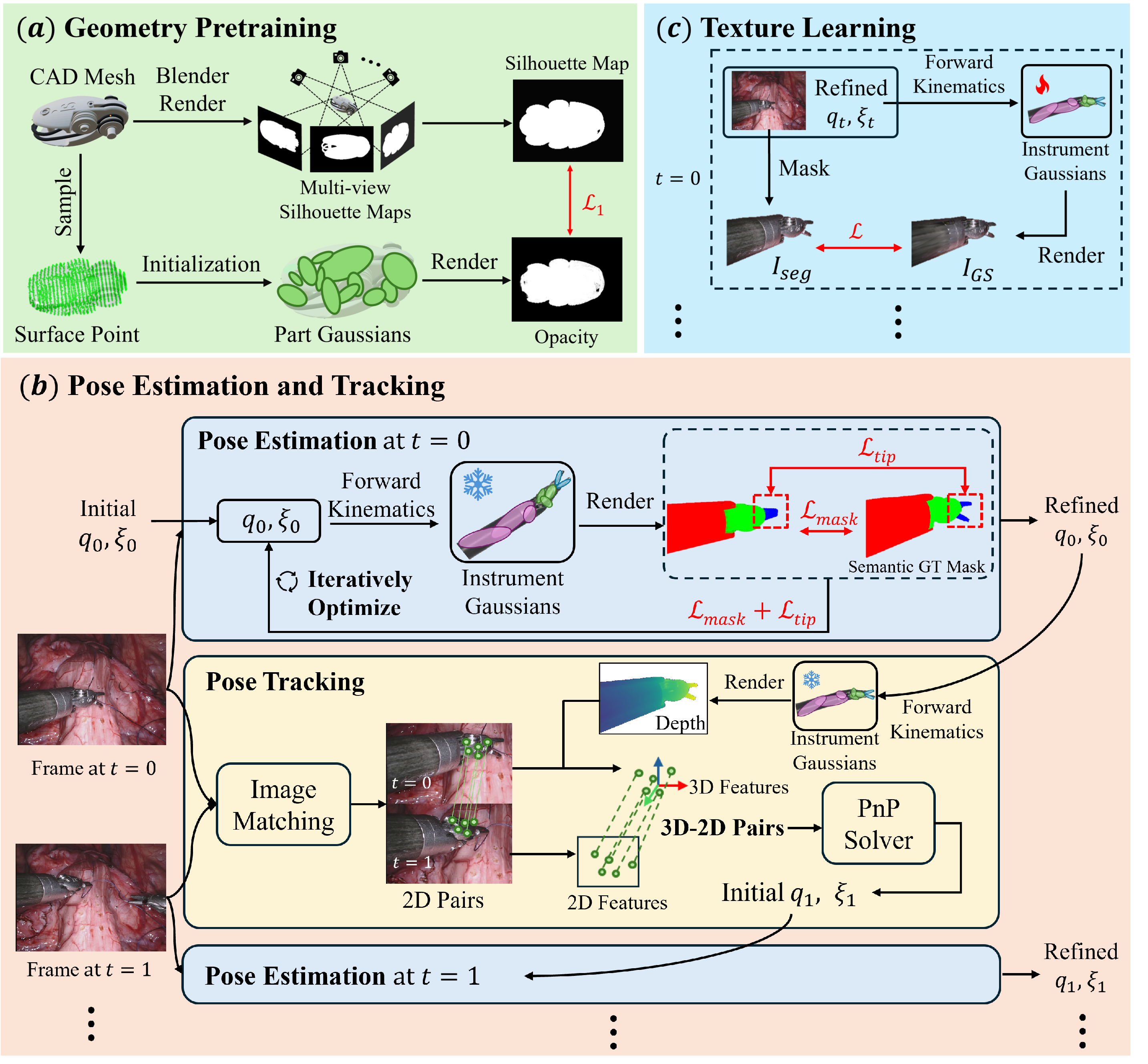}
\caption{Overview diagram of Instrument-Splatting methodology.} \label{fig3}
\end{figure}

\subsection{Geometry Pretraining} \label{2.2}
To bind the GS points to each rigid part of the instrument and initialize them with accurate geometric priors, we propose a geometry pretraining strategy, as illustrated in Fig.~\ref{fig3} (a). Specifically, for each part, we densely sample points on mesh surface by ray-tracing, which effectively avoids the sparse point distribution due to oversized triangles. Then, given the CAD model, we adopt Blender~\footnote{https://www.blender.org/} to render the ground-truth silhouette maps from multiple viewpoints. Next, the sampled GS points will be trained to learn the geometry information by minimizing the ${\mathcal L}_1$ loss between the rendered opacity and the ground-truth silhouette. Afterward, part Gaussians are combined to generate semantic-embedded instrument Gasussians where one semantic label (i.e., wrist, shaft, and gripper) $f \in \{1,2,3\}$ is added to Gaussian parameters $\Theta$ to identify the part the Gaussian point belongs to. Therefore, the instrument GS can be controlled by updating $\mu$ and $r$ with Eq.~\ref{eq:transformation} as:
\begin{equation}
    \mu'_j=\boldsymbol{T_j}(\mu_j;\;q,\xi);\;\;\;r'_j = \boldsymbol{R_j}(r_j;\;q,\xi),
\end{equation}
where $\mu'_j$ and $r'_j $ respectively denote the updated positions and rotations for Gaussian points in part $j$, $\boldsymbol R_j$ means the rotational component derived from rigid-body transformation $\boldsymbol{T_j}$.


\subsection{Pose Estimation and Tracking} \label{2.3}
An accurate alignment between the rendered instrument and the reference image is crucial for appearance learning. Therefore, we designe a pose estimation and tracking module to estimate per-frame instrument pose and joint changes, as illustrated in the Fig.~\ref{fig3} (b). 
Inspired by~\cite{allan20183}, the pose estimation adopts the render-and-compare framework to align the rendered semantic silhouette with the segmented instrument. 
For the render-and-compare method, a good pose initialization is crucial. In the initial frame, we manually pick 2D-3D point correspondences of logo landmarks in the wrist part and use the PnP to solve the pose. 
We cast the pose estimation task as an optimization problem that can be formulated as:
\begin{equation}
\widehat\xi,\;\widehat q=argmin\;\mathcal L(\xi,\;q)
\end{equation}
where the loss is $\mathcal L={\mathcal L}_{mask}+{\mathcal L}_{tip}$. The ${\mathcal L}_{mask}$ is ${\mathcal L}_{1}$ loss. To take advantage of the surgical instrument geometry prior, we propose ${\mathcal L}_{tip}={\mathcal L}_{dist}+{\mathcal L}_{struct}$ to enforce the tool tip alignment. Given the gripper mask, we use singular value decomposition (SVD) to find the principal axis of the gripper regions, on which the farthest pixel from the wrist part is regarded as the tool tip. We adopt a distance loss ${\mathcal L}_{dist}=ReLU(d - r)$, where $d$ is the Euclidean distance between the rendered tool tip from instrument GS and the estimated tool tip position from segmentation mask, and $r$ is a threshold to determine if the tool tip is aligned well. This loose regularization can alleviate falling into local minima caused by incorrect tool tip detection. In addition, we design a structural loss ${\mathcal L}_{struct} = ReLU(-(\theta_l+\theta_r))$ to constrain the left tip always on the left side of the right tip. $\theta_l$ is the left gripper rotation angle, which is a positive value if it rotates leftward and \textit{vice versa} it is a negative value. 

To track the instrument pose $\xi$ and joint state $q$, simply taking the previous frame pose as the initial value struggles to handle the case where the instrument motion between two adjacent frames is large. To address this issue, we propose a novel pose initialization method based on image matching and PnP for pose tracking. We extract features in the wrist and then find the matching correspondences~\cite{chen2022aspanformer} between two adjacent frames. Given the depth rendered from 3D GS, we can lift the 2D features to 3D. We approximate the frame $t$'s 3D features and frame $t+1$'s 2D features into 2D-3D correspondences at frame $t+1$ and then compute a rough initial pose using a PnP solver for the next render-and-compare refinement.

\subsection{Texture Learning} \label{2.4}
After estimating and tracking the pose and joint state over the video, we can control the instrument Gaussian points moving to the target pose in each frame, as shown in Fig.~\ref{fig3} (c). To maintain the geometric details, we freeze $\mu$ in this phase, and the other parameters $\{r,\;s,\;\alpha,\;sh\}$ are learnable during this section. Here, we adopt the loss ${\mathcal L}={\mathcal L}_{mask}+{\mathcal L}_{color}$ to learn the textures,
where the ${\mathcal L}_{mask}$ and ${\mathcal L}_{color}$ are the ${\mathcal L}_1$ loss of the semantic silhouette maps and RGB pixels of the instrument, respectively. ${\mathcal L}_{mask}$ is used here to maintain the geometric structure while learning appearances.

\section{Experiments and Results}
\subsection{Datasets and Implementation Details}
We establish a new benchmark for the evaluation of surgical instrument reconstruction. It includes two widely used instrument segmentation datasets  EndoVis 2017~\cite{allan20192017} and 2018~\cite{allan20202018}. 
We select 100 frames containing LND in both datasets. Frames with motion artifacts are excluded to ensure data quality. 
We split each video into seen and unseen views (7:1) for training and evaluation. 
Because of the sparse instrument poses in the \textit{in-vivo} datasets, we collect four videos with backgrounds of \textit{ex vivo} tissues and a green screen for fairer validation. 
Two trajectories with tissue background are to mimic the surgical scenario. 
The other two trajectories with green screen backgrounds are closer to the camera to capture more texture details. 
The data collection methodology follows SurgPose~\cite{wu2025surgpose}. We leverage a da Vinci IS1200 system with the da Vinci Research Kit (dVRK)~\cite{kazanzides2014open} to execute customized trajectories and record laparoscopic video stream. Regarding annotation, we use SAM 2~\cite{ravi2024sam} to segment the foreground surgical instrument at the part level and obtain corresponding semantic masks. 
This newly established benchmark with our collected in-house data will be publicly available soon.

In this work, we perform geometry pretraining, pose estimation, and texture learning sequentially. We first train the part Gaussian for 10K iterations per part. Then, we adopt 2K iterations to iteratively optimize the pose for each frame, and an early-stop signal will be triggered once the loss remains stable for over 100 iterations. The texture learning lasts for 10K iterations for each video.
During evaluation, we estimate the instrument poses and use these poses as user input to control the instrument GS by forward kinematics.
All the experiments are based on the PyTorch framework and conducted with a single NVIDIA RTX A5000 GPU.
\subsection{Quantitative Evaluation of Key Components}
\begin{figure}[h]
    \centering
    \includegraphics[width=0.8\textwidth]{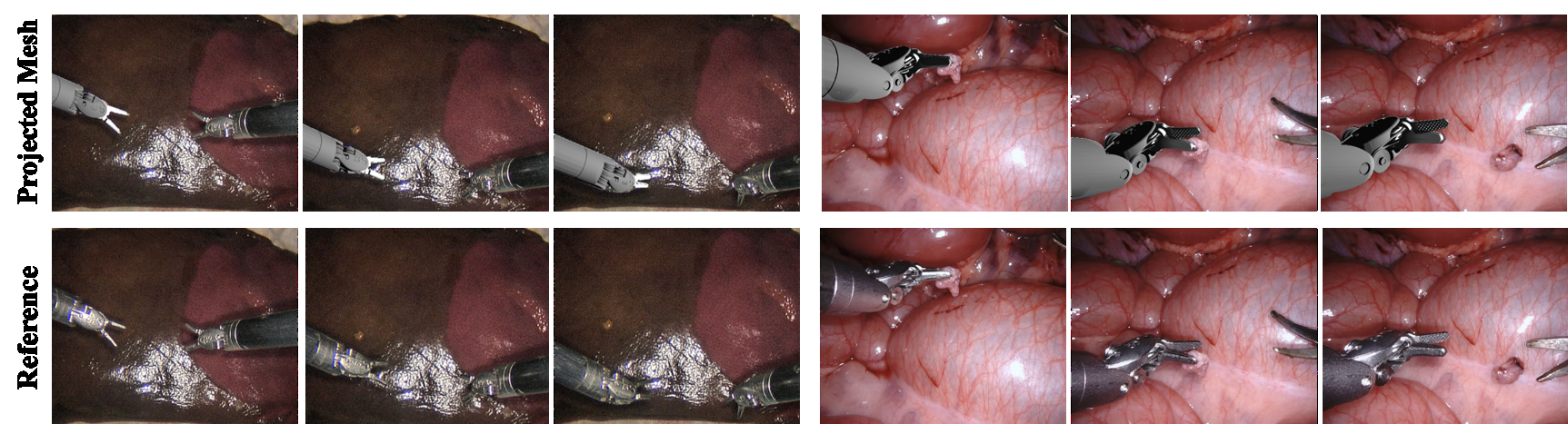}
    \caption{Visualization of the projected CAD mesh with the estimated poses and corresponding original images.}
    \label{pose_tracking}
\end{figure}

First, we verify the effectiveness of the key components of Instrument-Splatting on the EndoVis 2017 and 2018 datasets. As shown in Table~\ref{ablation}, we quantify the pose accuracy of our pose estimation and tracking module with 2D projection error similar to~\cite {allan20183}. Specifically, we adopt the difference (i.e., Dice score) between rendered semantics maps from posed instrument Gaussians and the corresponding segmentation maps to evaluate the pose accuracy. We test our method by removing the ${\mathcal L}_{tip}$ and the loose regularization strategy, denoted as $w/o\;{\mathcal L}_{tip}$ and $ w/o\; Reg$, respectively. As shown in Table~\ref{ablation}, both ${\mathcal L}_{tip}$ and the loose regularization lead to significantly improved pose estimation accuracy, particularly for the gripper. We also report the metrics of the reconstruction quality in Table~\ref{ablation}, where it is noticeable that an improved pose estimation can enhance the visual fidelity. Table~\ref{ablation} indicates the importance of Geometry Pretraining ($GP$), which significantly boosts the reconstruction quality. 
As shown in Fig.~\ref{pose_tracking}, six examples are selected from EndoVis18 and in-house dataset for visualization, which indicates that our method can robustly track the instrument pose. 

\begin{table}[t]
\caption{Quantitative Results of the Key Components}
\centering
\resizebox{0.8\textwidth}{!}{
\renewcommand{\arraystretch}{1} 
\begin{tabular}{>{\centering\arraybackslash}m{2cm} 
                >{\centering\arraybackslash}m{1.3cm} 
                >{\centering\arraybackslash}m{1.3cm} 
                >{\centering\arraybackslash}m{1.3cm} |
                >{\centering\arraybackslash}m{1.3cm} 
                >{\centering\arraybackslash}m{1.3cm} 
                >{\centering\arraybackslash}m{1.3cm}}
\hline
\multirow{2}{*}{Method} & \multicolumn{3}{c|}{Rendered Semantics Dice$\uparrow$} & \multicolumn{3}{c}{Reconstruction Quality} \\ \cline{2-7} 
                          & Shaft & Wrist & Gripper & PSNR$\uparrow$ & SSIM$\uparrow$ & LPIPS$\downarrow$ \\ \hline
$w/o\;{\mathcal L}_{tip}$ & 94.625 & 83.94 & 63.165 & 21.77 & 91.85 & 0.091 \\
$w/o\; Reg$               & 95.515 & 78.4 & 63.39 & 22.89 & 92.63 & 0.079 \\ \cline{1-7}
w/o GP                    & - & - & - & 14.6 & 89.54 & 0.16 \\
Ours                      & \textbf{96.83} &  \textbf{86.65} & \textbf{73.32} & \textbf{23.84} & \textbf{93.10} & \textbf{0.071} \\ \hline
\end{tabular}}

\label{ablation}
\end{table}
\subsection{Evaluation on Reconstruction Quality}
\begin{figure}[t]
    \centering
    \includegraphics[width=0.7\textwidth]{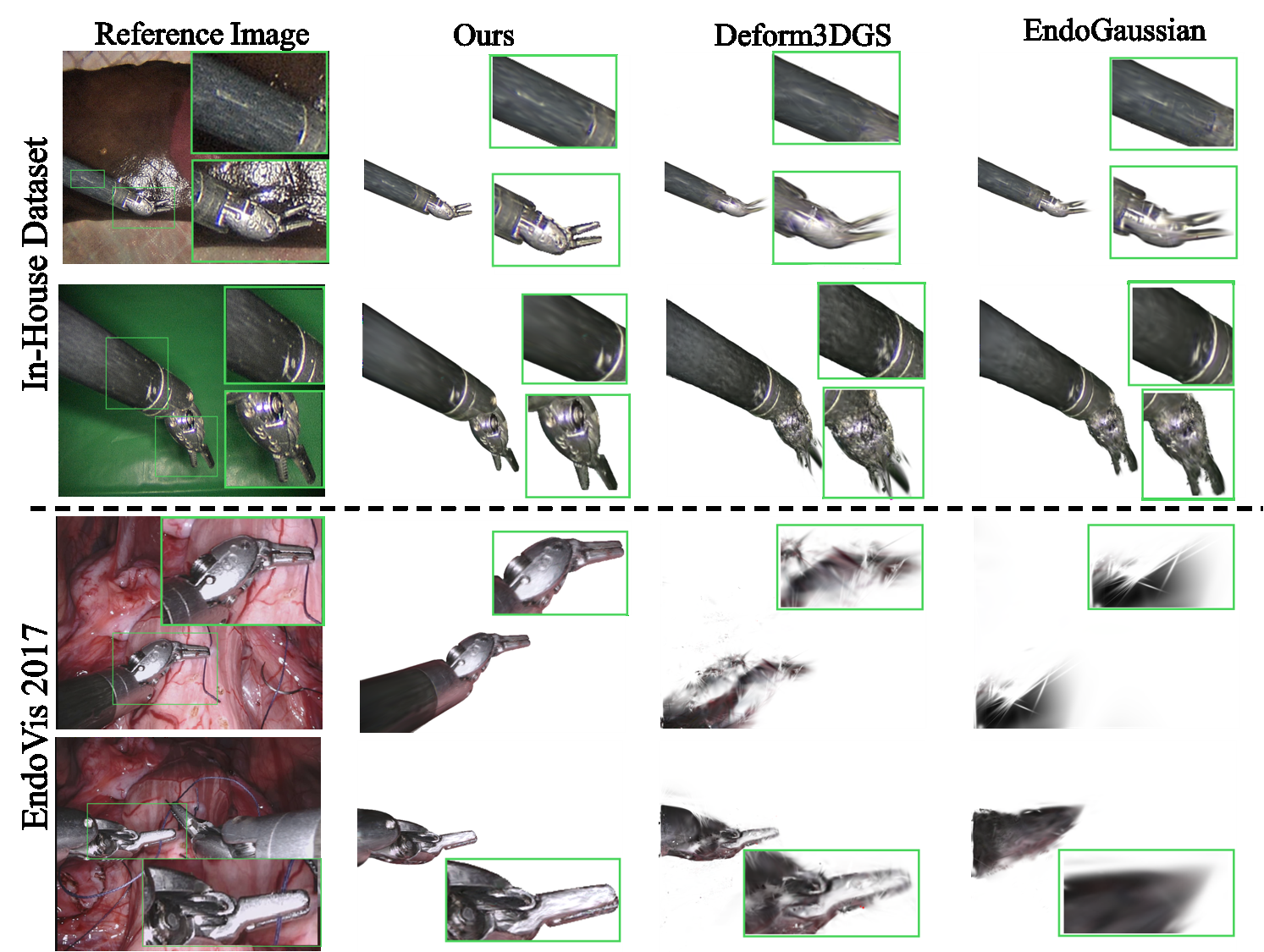}
    \caption{Visualization of the Novel-view Rendering Results of Different Methods}
    \label{visualization}
\end{figure}

In this work, we are targeting a novel setting that reconstructs articulated surgical instruments with monocular videos and CAD models only. We compare the reconstruction performance of our method with current state-of-the-art reconstruction methods in the surgical domain: EndoGaussian~\cite{liu2024endoG} and Deform3DGS~\cite{yang2024deform3dgs}. Although both methods cannot provide controllable 3D assets, they demonstrate high visual fidelity in the reconstruction of deformable tissues. 
As shown in Table~\ref{table2}, we show the results of wrist\&gripper, shaft, and the whole instrument separately. The metrics for the whole instrument are lower than the other two since they are computed based on all the pixels, following~\cite{liu2024endoG}. We visualize the rendered images of different methods in Fig.~\ref{visualization}. Both EndoGaussian and Deform3DGS fail to reconstruct instruments in real surgical videos (EndoVis2017\&2018) due to long-range motions and large time intervals. As a comparison, benefiting from the pose estimation module, our method can effectively capture the significant inter-frame pose change for real surgical videos, leading to a largely improved accuracy. On in-house data with smoother and slower motions, although Deform3DGS/EndoGaussian have comparable or even higher PSNR scores, their reconstruction outputs suffer from significant artifacts on wrist and gripper parts and are far from being photorealistic, as shown in the visualization (Fig.~\ref{visualization}). This can be explained by the motion patterns in some clips of our in-house data where the shaft part only undergoes limited translation, and Deform3DGS/EndoGaussian has a strong capability of capturing the appearance of these regions. Also, Table~\ref{table2} shows the two methods exhibit an inferior reconstruction quality on wrist and gripper parts, which indicates the existing methods on surgical scene reconstruction cannot handle articulated instrument parts with high dexterity and complex motions.
\begin{table}[]
\centering
\caption{Quantitative evaluation of the reconstruction quality on EndoVis2017\&2018 and In-House Dataset.}
\resizebox{\textwidth}{!}{\begin{tabular}{c|c|ccc|ccc|ccc}
\hline
\multirow{3}{*}{Dataset}       & \multirow{3}{*}{Method} & \multicolumn{3}{c|}{\multirow{2}{*}{Wrist \& Gripper}} & \multicolumn{3}{c|}{\multirow{2}{*}{Shaft}} & \multicolumn{3}{c}{\multirow{2}{*}{Overall Instrument}}  \\ 
                               &                         & \multicolumn{3}{c|}{}                                  & \multicolumn{3}{c|}{}                       & \multicolumn{3}{c}{}                            \\ \cline{3-11}
                               &                         & PSNR$\uparrow$           & SSIM$\uparrow$         & LPIPS$\downarrow$         & PSNR$\uparrow$         & SSIM$\uparrow$          & LPIPS$\downarrow$       & PSNR$\uparrow$           & SSIM$\uparrow$           & LPIPS$\downarrow$          \\ \hline
\multirow{3}{*}{EndoVis17\&18}      & Deform3DGS              & 19.65           & 92.55            & 0.165           & 22.23         & 94.33        & 0.066        & 17.44         & 86.49          & 0.213          \\
                               & EndoGaussian            & 18.90           & 93.45           & 0.145            & 20.91        & 94.30       & 0.078        & 16.45         & 87.75         & 0.192         \\
                               & Ours                    & \textbf{25.81 }         & \textbf{96.87  }          & \textbf{0.064}           & \textbf{31.39}       & \textbf{96.52}       & \textbf{0.035}       & \textbf{23.84}         & \textbf{93.10}           & \textbf{0.071}          \\ \hline
\multirow{3}{*}{In-House} & Deform3DGS              & 28.95            & 97.60            & 0.046            & \textbf{34.52}         & 97.08        & \textbf{0.023}        & \textbf{28.39}         & 94.76          & 0.048          \\
                               & EndoGaussian            & 25.34            & 96.94            & 0.059            & 31.59         & \textbf{97.23}        & 0.033        & 27.47         & 94.94          & 0.053          \\
                               & Ours                    & \textbf{30.44}            & \textbf{98.27}            & \textbf{0.038}            & 32.49         & 97.14        & 0.032        & 27.94         & \textbf{95.45}          & \textbf{0.041}          \\ \hline
\end{tabular}}
\label{table2}
\end{table}

\section{Conclusion}
In this paper, we propose Instrument-Splatting, a novel Real2Sim pipeline to reconstruct articulated robotic surgical instruments with high visual fidelity. With only monocular videos and CAD models of the instruments, our Instrument-Splatting can effectively track the per-frame instrument pose and learn the texture through RGB videos. Different from prior works in the surgical domain, our instrument Gaussian is articulated and fully controllable, which allows for manipulation as flexible as real instruments. We believe this digital twin of the instrument can strongly benefit the community working on surgical AI or surgical autonomy. The experimental results demonstrate the superior advances of our proposed method. The future work will focus on improving the current pose estimation module. A learning-based pose estimation probably has faster inference speed and better performance in terms of accuracy.

\bibliographystyle{SPLNCS04} 
\bibliography{Paper-3887.bib} 
\end{document}